\documentclass{article}
\usepackage{spconf}
\usepackage{amsmath,amssymb}
\usepackage{amsthm}
\usepackage{graphicx}
\usepackage{booktabs}
\usepackage{cite}
\usepackage{microtype}
\usepackage{array}
\usepackage{tikz}
\usepackage{multirow} 

\newcommand{\OC}{\operatorname{OC}}
\newcommand{\E}{\mathbb{E}}
\newtheorem{proposition}{Proposition}

\usepackage{etoolbox}
\usepackage[subtle]{savetrees}

\newcommand{\tightdisplayskips}{%
  \setlength{\abovedisplayskip}{5pt}%
  \setlength{\belowdisplayskip}{5pt}%
  \setlength{\abovedisplayshortskip}{5pt}%
  \setlength{\belowdisplayshortskip}{5pt}%
}

\AtBeginDocument{%
  \tightdisplayskips
  \apptocmd{\normalsize}{\tightdisplayskips}{}{}
  \apptocmd{\small}{\tightdisplayskips}{}{}
  \apptocmd{\footnotesize}{\tightdisplayskips}{}{}
}

\makeatletter

\renewcommand\section{\@startsection{section}{1}{\z@}%
  {9pt}{4pt}{\normalfont\bfseries}}

\renewcommand\subsection{\@startsection{subsection}{2}{\z@}%
  {6pt}{3pt}{\normalfont\bfseries}}

\renewcommand\subsubsection{\@startsection{subsubsection}{3}{\z@}%
  {6pt}{3pt}{\normalfont\itshape}}

\makeatother

\title{How Many Posterior Samples? Calibrated Stopping for Adaptive Sensing}
\name{Vincent Corlay and Andriy Enttsel}
\address{Mitsubishi Electric R\&D Centre Europe}

\begin{document}
\ninept
\maketitle

\begin{abstract}
In classification-oriented adaptive sensing, posterior samples characterize uncertainty at the current measurement state and can serve two roles: they may guide the next sensing direction, while their class labels provide votes for the candidate classes and determine whether sensing should continue. We focus on the stopping layer that turns these votes into a declaration, without modifying the posterior sampler or sensing directions. A natural plug-in rule declares when the observed vote share exceeds a threshold. We show that this threshold is not itself a confidence guarantee: when the underlying vote mass equals the threshold, the plug-in rule declares about half the time. As alternatives, we calibrate a fixed-sample rule and a finite-horizon sequential rule to a prescribed false-declaration probability, and study \emph{exact curtailment}, which stops a fixed-pool rule once its final verdict is forced. We then derive how one-round declaration probabilities determine posterior-sample cost and classification accuracy along a sensing path. On MNIST with DDRM and a fixed PCA-guided probe sequence, curtailment saves up to 62\% of posterior samples. Among the evaluated rules at matched operating points, sequential stopping reduces the cost the most. At a high accuracy, that same sequential rule can trade more posterior samples for fewer measurements.
\end{abstract}

\begin{keywords}
adaptive sensing, diffusion posterior sampling, sequential testing, sampling budget, active classification
\end{keywords}

\section{Introduction}

\begin{figure}[t]
\centering
\resizebox{\columnwidth}{!}{%
\begin{tikzpicture}[
    font=\footnotesize,
    >=stealth,
    block/.style={
        draw=black,
        line width=0.7pt,
        rounded corners=1.5pt,
        align=center,
        minimum height=7.5mm,
        inner xsep=2.5pt,
        inner ysep=1.5pt
    },
    flow/.style={
        ->,
        line width=0.8pt
    },
    outer/.style={
        ->,
        line width=1.0pt,
        red!75!black,
        shorten >=1pt
    },
    inner/.style={
        ->,
        line width=1.0pt,
        blue!70!black,
        shorten >=1pt
    }
]

\node[block, text width=17mm] (measurement) at (0,0)
    {measurement\\state $y_t$};

\node[block, text width=18mm] (sampler) at (2.45,0)
    {posterior\\sampler};

\node[block, text width=15mm] (classifier) at (4.65,0)
    {classifier};

\node[block, text width=18mm] (vote) at (6.90,0)
    {vote state\\$(n,k)$};

\node[anchor=west] (declare) at (8.35,0)
    {declare $\hat c$};

\draw[flow] (measurement.east) -- (sampler.west);
\draw[flow] (sampler.east) -- (classifier.west);
\draw[flow] (classifier.east) -- (vote.west);
\draw[flow] (vote.east) -- (declare.west);

\draw[outer]
    (vote.north)
    -- (6.90,0.8)
    -- (0,0.8)
    -- (measurement.north);

\node[
    text=red!75!black,
    align=center,
    font=\scriptsize
] at (3.45,1.1)
    {keep sensing\\[-1pt]
     outer budget: one measurement round};

\draw[inner]
    (vote.south)
    -- (6.90,-0.8)
    -- (2.45,-0.8)
    -- (sampler.south);

\node[
    text=blue!70!black,
    align=center,
    font=\scriptsize
] at (4.65,-1.1)
    {sample again\\[-1pt]
     inner budget: one reverse-diffusion trajectory};

\end{tikzpicture}%
}
\vspace{-7mm}
\caption{At each measurement state, posterior samples are classified into
votes. The vote rule may request another posterior sample, declare a class,
or return control to the outer sensing loop.}
\label{fig:adaptive_loop}
\end{figure}
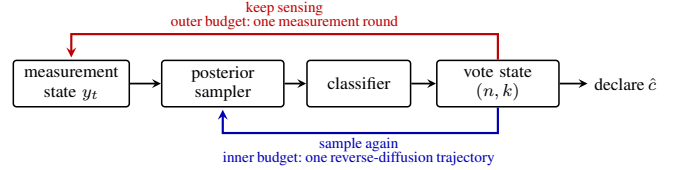

Diffusion models can generate plausible signals compatible with measurements observed so far~\cite{ddrm,dnsm}. In adaptive sensing, such posterior samples may guide the next probe~\cite{adaptivecs} and, for classification, may also determine whether another measurement is needed~\cite{taskuq, classpost}. This creates two coupled loops, illustrated in Figure~\ref{fig:adaptive_loop}: an inner loop generates and classifies posterior samples at the current measurement state, while the outer loop acquires another measurement if the inner loop does not declare. Since each posterior sample requires a reverse-diffusion trajectory, the rule deciding when enough samples have been seen can have a substantial computational effect.

To make this stopping problem concrete, suppose that at one measurement state we draw $S=32$ posterior images. If 23 receive the same digit label, their observed vote share is $23/32\approx0.72$, and a natural plug-in rule with threshold $\tau=0.70$ would declare that digit. What does such a declaration mean statistically? Although $\tau=0.70$ may look like a confidence level, it is only a threshold on the observed vote share. If the underlying probability of voting for that digit were exactly $0.70$, a 32-sample pool would still exceed the threshold and trigger a declaration about half the time. 
Thus, the vote threshold alone does not quantify the reliability of the declaration.

This agreement-based stopping approach is not unique to adaptive sensing. Related ideas appear in adaptive-consistency methods for language models~\cite{adaptiveconsistency,bayesstop} and, more generally, in classical sequential testing~\cite{wald,safe}. Our contribution is not agreement-based stopping itself, but its calibration and cost analysis when embedded in a sensing loop. In this setting, the same vote statistic controls two resources: posterior sampling at the current measurement state and, through the decision to declare or not, whether another physical measurement is acquired. Moreover, the class receiving the strongest vote support need not be the ground-truth class.

This leads to three related but distinct questions. \emph{Certification} asks whether a class with insufficient underlying vote support is rarely declared. \emph{Accuracy} asks whether the class eventually declared by the full sensing process is the true class. \emph{Cost} counts the posterior samples generated along the way. A stopping rule may perform well on one criterion without controlling the others.

\noindent\textbf{Contributions.} We study the stopping and calibration layer of this classification loop, without modifying the posterior sampler or sensing directions. First, we show why the plug-in vote threshold is not a confidence level and propose a calibrated counterpart (the one-look rule). Second, we calibrate a finite-horizon sequential rule through the probability of declaring before the maximum inner budget, enabling fixed and sequential rules to be compared at matched attained false-declaration probability (type I error). Third, we formalize exact curtailment and quantify its sample savings without changing any terminal decision. Fourth, we derive how the one-round declaration curve determines both posterior-sample cost and classification accuracy along a sequence of measurement states. Finally, we validate these predictions on an MNIST/DDRM \cite{ddrm} sensing loop.

\vspace{-2mm}

\section{Vote model and stopping rules}\label{sec:votes}

Fix the current measurements $y$. Each reverse-diffusion draw $x\sim p_\theta(\cdot\mid y)$ is classified into one of $C$ classes and contributes one hard vote. Repeating this experiment indefinitely at fixed $y$ induces the categorical distribution over vote labels
\begin{align}
\boldsymbol\pi(y)&=(\pi_1(y),\ldots,\pi_C(y)), \quad \text{with}\\
\pi_c(y)&=\Pr(\text{the next posterior sample votes for }c\mid y).
\end{align}
For a class $c$, write $q=\pi_c(y)$ and call it the \emph{population vote mass}. It is the long-run vote fraction for $c$ at this measurement state, not the probability that $c$ is the true class, and it reflects the combined behavior of the posterior sampler and classifier.

The quantity $q$ is hidden. After $n$ draws, we observe only the vote counts: let $k_{n,c}$ denote the number of votes by the classifier for class $c$, with empirical share $k_{n,c}/n$. The considered stopping rules use these counts to decide whether to declare a class.

We call the posterior samples generated at one fixed measurement state a \emph{pool}, and their number the \emph{pool size}. A fixed-pool rule plans to draw $S$ samples before its terminal decision, although curtailment may stop earlier once that decision is already forced. If no class is declared, another physical measurement is acquired, and a new pool begins. Let $D$ denote the event that a class is declared, and $\hat c$ the declared class.

Across outer sensing rounds, the measurement state changes. At round $t$, let $\boldsymbol\pi_t=\boldsymbol\pi(y_t)$. The sequence $(\boldsymbol\pi_1,\boldsymbol\pi_2,\ldots)$ is the \emph{sensing path} in vote space and describes how the vote distribution evolves as measurements accumulate. The full path matters for loop-level cost and accuracy. For calibration, however, it is enough to consider one class at one fixed state: for the threshold rules considered here, whether class $c$ reaches the declaration criterion depends only on its own vote count, not on how the remaining votes are split among other classes. Under the conditional i.i.d. vote model, if class $c$ has vote mass $q=\pi_{t,c}$, then $k_{n,c}\sim\mathrm{Bin}(n,q)$. The predicted-versus-measured path cost and accuracy in Sec.~\ref{sec:experiments} provide an indirect consistency check of this model (but they do not isolate the i.i.d. assumption from error in estimating $\boldsymbol\pi_t$).

\subsection{Vote-based stopping rules}

Table~\ref{tab:rules} summarizes the four rules considered. The plug-in and one-look rules make their terminal decision after a prescribed number $S$ of samples. Sequential stopping and curtailment can stop earlier, but for different reasons.

\begin{table}[t]
\centering
\caption{\footnotesize Vote-based stopping rules and their use of early stopping.}
\label{tab:rules}
\footnotesize
\setlength{\tabcolsep}{2.5pt}
\renewcommand{\arraystretch}{1.06}
\begin{tabular}{@{}>{\raggedright\arraybackslash}p{0.2\columnwidth}>{\raggedright\arraybackslash}p{0.74\columnwidth}@{}}
\toprule
Rule & What the rule does \\
\midrule
Plug-in & Wait for $S$ samples, then declare the empirical leader if its observed share exceeds $\tau$. \\
One-look & Wait for $S$ samples, then declare only if the leader reaches a count calibrated to a chosen error probability. \\
Sequential & Inspect the count after every sample and declare when a calibrated sequential boundary is first crossed. \\
Curtailment & Apply a fixed-pool rule, but stop generating samples once every possible continuation gives the same final verdict. \\
\bottomrule
\end{tabular}
\end{table}

\noindent\textbf{Plug-in rule.} Let $k_{S,(1)}=\max_c k_{S,c}$ be the largest class count after $S$ samples. 
The rule declares iff $k_{S,(1)}/S>\tau$, equivalently iff $k_{S,(1)}\ge r$, where $r=\lfloor S \times \tau\rfloor+1$. 
In the running example, $S=32$ and $\tau=0.70$, so $r=23$. 
Thus the plug-in rule is already a \textit{critical-count rule}. However, as Sec.~\ref{sec:calibration} shows, it does not directly control the false-declaration probability.

\noindent\textbf{Calibrated one-look rule.} Suppose instead that we require any class with vote mass $q\le\tau$ to be declared with probability at most $\varepsilon$. 
Since $k_{S,c}\sim\mathrm{Bin}(S,q)$, declaring class $c$ at the critical
count $r$ occurs with probability $\Pr_q(k_{S,c}\ge r)$.
This probability increases with $q$. The worst case over $q\le\tau$ therefore occurs at the boundary $q=\tau$. We therefore choose the smallest count threshold whose binomial tail at $q=\tau$ is at most $\varepsilon$:
\begin{equation}
r=\min\{j:\Pr(K\ge j)\le\varepsilon,\ K\sim\mathrm{Bin}(S,\tau)\}.
\label{eq:one-look}
\end{equation}
For $S=32$, $\tau=0.70$, and $\varepsilon=0.05$, this gives $r=28$. The attained false-declaration probability is $0.019$, rather than exactly $0.05$, because the binomial count is discrete. The rule still waits for all 32 samples, but its critical count now has a direct statistical interpretation.

\noindent\textbf{Sequential rule.} Waiting for all $S$ samples can be wasteful when the evidence becomes compelling earlier. After $n$ samples, we declare the class $c$ if the state $(n,k_{n,c})$ satisfies
\begin{equation}
\Pr(\pi_c>\tau\mid k_{n,c},n)>1-\alpha.
\label{eq:seq}
\end{equation}
Under the Bernoulli vote model and a uniform prior on $\pi_c$, observing $k_{n,c}$ votes out of $n$ gives $\pi_c\mid k_{n,c},n\sim\mathrm{Beta}(k_{n,c}+1,n-k_{n,c}+1)$. We evaluate \eqref{eq:seq} using this posterior and impose a finite cap (i.e., per round inner budget limit) $n_{\max}$. If no continuation can reach the declaration region by the cap, the rule returns ``keep sensing''. This exact abandonment does not change whether the class would ever be declared by $n_{\max}$.

The inequality \eqref{eq:seq} requires the posterior to exceed $1-\alpha$, which occurs only when $k_{n,c}$ is large enough. It therefore defines a count boundary $b(n)$ on the $(n,k)$ lattice: class $c$ is declared when $k_{n,c}\ge b(n)$. The parameter $\alpha$ is \emph{not} the overall frequentist false-declaration probability because the rule has repeated opportunities to cross that boundary. We therefore calibrate that declaration probability in Sec.~\ref{sec:calibration}.

\noindent\textbf{Exact curtailment.} A fixed-pool rule can be stopped early without changing its terminal decision. For the one-look running example, declaration requires $r=28$ votes in a pool of size $S=32$. If the $28$th vote for class $c$ is observed before the pool is complete, declaration is already determined. Conversely, if five votes have already been assigned to other classes, the largest attainable count for $c$ is $27$, so nondeclaration is forced.
More generally, consider any fixed-pool rule (e.g., plug-in) with critical count $r$. After $n$ draws, declaration is forced when $k_{n,c}\ge r$, while nondeclaration is forced when $k_{n,c}+(S-n)<r$, because the left-hand side is the largest final count that class $c$ can still reach. Exact curtailment stops as soon as either outcome is logically determined~\cite{curtailment}. It saves samples while returning the same terminal verdict as the full fixed pool on every vote sequence.

\section{What does a certificate mean?}\label{sec:calibration}

A certificate should answer a repeated-sampling question: if a class has limited underlying vote support, how often would the stopping rule nevertheless declare it? Consider a class $c$ at one measurement state and let its population vote mass be $q$. Repeating the posterior-sampling experiment at that state gives the declaration probability
\begin{equation}
\OC(q)=\Pr(\hat c=c,D\mid \pi_c=q),
\label{eq:oc}
\end{equation}
the operating characteristic of the rule~\cite{wald}.

For the monotone count rules considered here, declaration becomes more likely as $q$ increases. Hence the largest declaration probability among classes with $q\le\tau$ occurs at the boundary:
\begin{equation}
\varepsilon_{\rm class}
=\sup_{q\le\tau}\OC(q)
=\OC(\tau).
\label{eq:size}
\end{equation}
We call $\varepsilon_{\rm class}$ the classwise false-declaration probability (also often called the size or Type~I error). A rule is \emph{calibrated} when this error level can be controlled by design. Thus, a ``5\% certificate at $\tau$'' means that any class with vote mass $q\le\tau$ is declared with probability at most $0.05$.

The plug-in rule does not provide such control: it uses $\tau$ as an empirical vote threshold, but $\tau$ does not determine $\varepsilon_{\rm class}$. Proposition~\ref{prop: plug-in} makes this failure explicit.

\begin{proposition}[Plug-in boundary value]
\label{prop: plug-in}
Let $r_S=\lfloor S \times \tau\rfloor+1$ and assume $\tau\ge 1/2$. At the boundary $q=\tau$,
\begin{equation}
\OC(\tau)=\Pr(K\ge r_S)
=\tfrac12+O(S^{-1/2}),\quad
K\sim\mathrm{Bin}(S,\tau),
\label{eq:plugin}
\end{equation}
with a remainder that oscillates with the fractional part of $S \times \tau$.
\end{proposition}

Proposition~\ref{prop: plug-in} formalizes the running example. When the population vote mass equals the threshold, the empirical vote share is centered at that same value. Increasing $S$ makes the share concentrate more tightly around $\tau$, but does not drive the false-declaration probability $\OC(\tau)$ to zero: $\OC(\tau)$ remains near one half. For $S=8,16,32,64$, the corresponding values are $0.55$, $0.45$, $0.50$, and $0.54$, and even at $S=1024$ it is $0.51$.

The one-look rule avoids this problem by choosing its critical count directly from the binomial tail in \eqref{eq:one-look}. Because it makes only one decision after all $S$ votes are observed, that tail probability directly controls its false-declaration probability. The sequential rule is different: it checks after every new sample, and the posterior threshold $\alpha$ in \eqref{eq:seq} is not itself the overall false-declaration probability.

We therefore calibrate the sequential rule by choosing $(\alpha,n_{\max})$ so that $\OC(\tau)\le\varepsilon$ in \eqref{eq:size}. The recursion in Appendix~\ref{sec:firsthit} is used to choose $(\alpha,n_{\max})$. In testing terms, the calibration distinguishes $q\le\tau$ from $q>\tau$: the design controls the false-declaration probability $\OC(\tau)$ while achieving a high declaration probability $\OC(q_{\mathrm{alt}})$ at a reference alternative $q_{\mathrm{alt}}>\tau$, which we call the power. For example, with $\tau=0.70$ and $q_{\mathrm{alt}}=0.85$, the choice $n_{\max}=97$ and $\alpha=0.0091$ gives $\OC(0.70)=0.049$ and $\OC(0.85)=0.925$.


\section{From one round to the full sensing loop}

\subsection{Saving samples without changing a decision}

Curtailment changes only how many samples are generated, not what the fixed-pool rule ultimately decides. Let $N$ denote the number of posterior samples generated at one measurement state before the inner rule returns either ``declare'' or ``keep sensing.'' For a full fixed pool, $N=S$ deterministically. Under curtailment or sequential stopping, $N$ may be smaller.

\begin{proposition}[Exact curtailment]
Consider any rule whose terminal decision is defined after a fixed budget $S$. Stop at the first $n\le S$ for which every assignment of the remaining $S-n$ samples gives the same terminal decision. The curtailed and full rules then agree on every sample sequence. Consequently, all declaration probabilities, error rates, and the complete operating characteristic are unchanged.
\end{proposition}

For a critical-count rule, the $(n+1)$st sample is drawn if and only if the decision is still open after $n$ draws, that is, iff $k_{n,c}<r\le k_{n,c}+S-n$. The expected curtailed pool size is therefore
\begin{equation}
\E_q[N]=\sum_{n=0}^{S-1}
\Pr\nolimits_q\bigl\{k_{n,c}<r\le k_{n,c}+S-n\bigr\}.
\label{eq:curtail}
\end{equation}
Each summand is the probability that the fixed-pool verdict remains undecided after $n$ samples and another posterior sample is therefore required.

Curtailment is thus decision-preserving. A sequential rule can reduce sample cost further by changing the stopping rule itself, but the comparison is meaningful only after matching attained false-declaration probability and power. At $\tau=0.70$ and $q_{\mathrm{alt}}=0.85$, the smallest fixed-pool design matching the sequential rule's attained false-declaration probability and power has $S^\star=69$. After curtailment, the matched fixed-pool design has an expected cost of 64 samples, versus 45 for the sequential design; the latter is close to the information lower bound in Appendix~\ref{sec:lower}. As shown there, this advantage is regime-dependent rather than universal.

\subsection{Accumulating cost over measurement rounds}

The previous subsection gives the expected sample cost at one measurement state. Over a full sensing run, that cost is incurred only if the corresponding round is reached. Let $\rho_t$ denote the probability of reaching round $t$. Conditional on a prescribed sensing path $(\boldsymbol\pi_t)$ and with independently restarted pools, reaching round $t$ means that none of the earlier rounds declared:
\begin{equation}
\rho_t=\prod_{s<t}\bigl(1-\Pr(D\mid\boldsymbol\pi_s)\bigr).
\label{eq:reach}
\end{equation}
The expected posterior-sample cost along the path is therefore
\begin{equation}
\E[N_{\rm path}\mid(\boldsymbol\pi_t)]
=\sum_t\rho_t\,\E[N\mid\boldsymbol\pi_t].
\label{eq:pathcost}
\end{equation}
Thus, each one-round cost is weighted by the probability that the sensing loop reaches that round.

\subsection{Certification is not classification accuracy}

Certification concerns whether a declared class has sufficient \emph{vote support}. It does not guarantee that the class is the ground truth $c_{\rm true}$. A calibrated rule may therefore favor the wrong class if the posterior sampler and classifier assign most of the vote mass to it. If $c^\star$ is the vote-law leader with mass $\pi_{(1)}$ and $\tau\ge1/2$, then
\begin{equation}
\Pr(\hat c=c^\star\mid D,\boldsymbol\pi)\ge
\frac{\OC(\pi_{(1)})}
{\OC(\pi_{(1)})+\OC(1-\pi_{(1)})}.
\end{equation}
Appendix~\ref{sec:leader-bound} gives the proof. At the operating points considered here, the bound is essentially one: the rule can reliably identify the vote-law leader even when that leader is not the true class.

End-to-end accuracy must therefore account for both \emph{when} the loop first declares and \emph{which class} it declares. Conditional on a sensing path, let $d_t=\Pr(D\mid\boldsymbol\pi_t)$ and $\lambda_t=\rho_t d_t/\sum_s\rho_s d_s$, the probability that round $t$ is the first declaring round given that a declaration occurs. At that round, the probability of declaring the true class is
\begin{equation}
a_t=\frac{\OC(\pi_{t,c_{\rm true}})}{\sum_c\OC(\pi_{t,c})}.
\end{equation}
The path accuracy is therefore $A((\boldsymbol\pi_t))=\sum_t\lambda_t a_t$. Averaging across sensing paths gives
\begin{equation}
\Pr(\hat c=c_{\rm true}\mid D)=
\frac{\E_{(\boldsymbol\pi_t)}[A((\boldsymbol\pi_t))\Pr(D\mid(\boldsymbol\pi_t))]}
{\E_{(\boldsymbol\pi_t)}[\Pr(D\mid(\boldsymbol\pi_t))]}.
\label{eq:loopacc}
\end{equation}
Thus, end-to-end accuracy is obtained from the same one-round operating characteristic $\OC$, evaluated at the class vote masses encountered along the sensing path. In the experiments, the unknown $\boldsymbol\pi_t$ is estimated from logged posterior votes.

\section{Diffusion-based sensing experiments}\label{sec:experiments}

We evaluate the stopping rules on 1000 MNIST images using a fixed sequence of principal component analysis (PCA)-guided measurement directions, a pretrained DDRM sampler~\cite{ddrm} with 20 reverse steps, and a classifier. The PCA directions are fixed throughout the run and are not recomputed from posterior samples inside the stopping loop. This differs from adaptive variants that re-estimate a covariance from posterior samples and redesign the next probe~\cite{adaptivecs}. Here, the fixed probe sequence isolates the effect of the stopping rule.

One vote consists of one diffusion generation followed by classification. At each measurement state we log $S=32$ votes, which also sets $n_{\max}=32$ for the sequential rule. The $n_{\max}=128$ sequential row is replayed on a 128-vote log at the same measurement states. Each image is first run to the full sensing budget, and the four rules are then replayed on the same log. Thus, every rule sees the same measurement sequence $(y_t)$ and underlying vote-law path $(\boldsymbol\pi_t)$. Only the number of posterior samples used for stopping and the round at which a declaration occurs can differ.

Each block of Table~\ref{tab:mnist} reports the rules at nearly the same measured accuracy. The rows can therefore be compared on sample cost. The gap between ``plug-in (full)'' and ``plug-in (curt.)'' is the saving from curtailment, and the gap between one-look and sequential compares two different stopping rules.
The column $\tau/\tau'$ separates the design threshold $\tau$ from the certified vote mass $\tau'$, defined for $\epsilon_{class}=$5\% in \eqref{eq:size}. Accuracy is reported as measured (M) and as predicted from \eqref{eq:loopacc} using $\boldsymbol\pi_t$ estimated from the 32-vote rule log (P$_{32}$), and from a separate 128-vote estimate at the same measurement states (P$_{128}$). The 128-vote estimate is used only to assess error in estimating $\boldsymbol\pi_t$ and is not charged to the stopping rule. Posterior-sample cost is also reported as measured (M) and as predicted from \eqref{eq:pathcost} using $\boldsymbol\pi_t$ from the 32-vote rule log (P$_{32}$) and from the same 128-vote estimate (P$_{128}$).




\begin{table}[t]
\centering
\caption{\footnotesize MNIST sensing results with all considered rules at three target accuracies.}
\label{tab:mnist}
\footnotesize
\setlength{\tabcolsep}{2.6pt}
\renewcommand{\arraystretch}{1.06}
\begin{tabular}{@{}lcccc@{}}
\toprule
\multirow{2}{*}{Scheme}
& \multirow{2}{*}{$\tau/\tau'$}
& Accuracy & Rounds & Samples \\
& & M/P$_{32}$/P$_{128}$ & M & M/P$_{32}$/P$_{128}$ \\
\midrule

\multicolumn{5}{c}{\itshape Target measured accuracy $\approx 1.00$}\\[-1pt]
Plug-in (full)  & .90/.77 & .98/.97/.98 & 12.3 & 406/384/399 \\
Plug-in (curt.) & .90/.77 & .98/.97/.98 & 12.3 & 154/137/144 \\
One-look        & .75/.77 & .98/.97/.98 & 12.3 & 154/137/144 \\
Sequential ($n_{\max}{=}32$)  & .78/.78 & .98/.98/.98 & 12.5 & \textbf{118}/105/110 \\
Sequential ($n_{\max}{=}128$) & .78/.78 & .98/.97/.98 & \textbf{11.8} & 542/487/521 \\
\addlinespace[3pt]

\multicolumn{5}{c}{\itshape Target measured accuracy $\approx 0.95$}\\[-1pt]
Plug-in (full)  & .80/.66 & .96/.94/.95 & 10.5 & 339/320/333 \\
Plug-in (curt.) & .80/.66 & .96/.94/.95 & 10.5 & 173/156/162 \\
One-look        & .65/.66 & .96/.94/.95 & 10.5 & 173/156/162 \\
Sequential      & .65/.65 & .96/.93/.95 & 10.5 & \textbf{142}/125/130 \\
\addlinespace[3pt]

\multicolumn{5}{c}{\itshape Target measured accuracy $\approx 0.90$}\\[-1pt]
Plug-in (full)  & .70/.56 & .92/.88/.91 & 8.9 & 285/263/276 \\
Plug-in (curt.) & .70/.56 & .92/.88/.91 & 8.9 & 176/156/163 \\
One-look        & .55/.56 & .92/.88/.91 & 8.9 & 176/156/163 \\
Sequential      & .55/.55 & .91/.87/.90 & 8.8 & \textbf{149}/131/138 \\
\bottomrule
\end{tabular}
\end{table}

The first pair of rows shows the largest decision-preserving saving. At the highest target, curtailment reduces the measured cost from 406 to 154 posterior samples, a 62\% reduction, while leaving accuracy and the 12.3 measurement rounds unchanged. The corresponding costs at the other two targets fall from 339 to 173 and from 285 to 176 samples.

The plug-in and one-look rows also illustrate why a design threshold should not be read as a certificate. At the highest target, the plug-in rule uses $\tau=0.90$ and the one-look rule $\tau=0.75$, yet both correspond to the same critical count and certify only up to $\tau'=0.77$. The sequential design uses $\tau=0.78$ and attains $\tau'=0.78$. Thus, the vote threshold, certified vote mass, and measured classification accuracy are distinct quantities.

Among the rules compared at nearly the same measured accuracy, sequential stopping uses the fewest posterior samples in all three blocks: 118, 142, and 149, versus 154, 173, and 176 for one-look.

Among the $S/n_{\max}=32$ rules, the highest-accuracy block also uses the fewest posterior samples, despite more measurement rounds. This can be understood from the path-cost decomposition \eqref{eq:pathcost}: the higher accuracy target uses a higher declaration threshold, so on early states of the sensing path the vote mass is still too low to declare, and curtailment and sequential abandonment return ``keep sensing'' after a short pool. Only in the last rounds, when the vote mass is more concentrated, does the inner loop incur a larger cost.
At a high accuracy, a larger sequential inner-budget limit can trade more posterior samples ($118$ to $542$) for fewer measurement rounds ($12.5$ to $11.8$).

The path formulas predict the measured quantities reasonably well: sample-cost predictions from the 32-vote estimate run 5--12\% below the measured budgets, about 10\% on average, and tighten to about 2--9\% (about 6\% on average) with the 128-vote estimate. It likewise improves the accuracy predictions obtained from 32 votes. This agreement is consistent with the conditional i.i.d. vote model, but the remaining gap is not isolated from residual error in $\boldsymbol\pi_t$.

\section{Conclusion}

Adaptive sensing with diffusion posterior sampling requires deciding when the current posterior evidence is sufficient to stop acquisition. We showed that an empirical vote threshold is not itself a confidence guarantee, and developed calibrated fixed-sample and sequential stopping rules together with exact curtailment, which reduces computation without changing fixed-pool decisions. By linking the one-round operating characteristic to sensing-loop sample cost and classification accuracy, we separate certification of vote support from end-to-end task accuracy. On MNIST/DDRM, curtailment removes a large fraction of unnecessary posterior generations, while sequential stopping reduces cost further at nearly the same measured accuracy. At high accuracy, sequential stopping can trade more posterior samples for fewer measurements. A natural extension is to exploit dependence across successive measurement states to reuse or update posterior samples across sensing rounds.
These results show that substantial computational savings are available at the stopping layer alone, without modifying the posterior sampler or sensing directions.

\section{Appendix}


\subsection{Recursion for sequential calibration}\label{sec:firsthit}
For one specified class, each vote is Bernoulli with success probability $q$. The sequential state is $(n,k)$: the next vote moves to $(n+1,k+1)$ with probability $q$ and to $(n+1,k)$ with probability $1-q$. Let $u(n,k)$ be the probability of eventual declaration from state $(n,k)$, and let $e(n,k)$ be the expected final stopping count $N$. Declaration states (those at which the criterion of \eqref{eq:seq} is met) have $u=1$ and $e=n$. Cap and abandonment states have $u=0$ and $e=n$. At interior states,
\begin{align*}
u(n,k)&=q\,u(n+1,k+1)+(1-q)u(n+1,k),\\
e(n,k)&=q\,e(n+1,k+1)+(1-q)e(n+1,k).
\end{align*}
Evaluating the recursion from the initial state $(n,k)=(0,0)$ gives $\OC(q)=u(0,0)$ and $\E_q[N]=e(0,0)$. The same calculation therefore supplies both calibration and expected cost.

\subsection{Information lower bound}\label{sec:lower}
\begin{proposition}[Information lower bound]
Fix $\tau<q$. Let $N\le n_{\max}$ be the stopping time of any sequential rule on i.i.d. Bernoulli votes, and let $A\in\mathcal F_N$ be its declaration event. If $\Pr_\tau(A)\le\varepsilon$ and $\Pr_q(A)\ge1-\beta>\varepsilon$, then
\[
\E_q[N]\ge\frac{d(1-\beta\Vert\varepsilon)}{d(q\Vert\tau)},\qquad
d(x\Vert y)=x\log\frac{x}{y}+(1-x)\log\frac{1-x}{1-y}.
\]
\end{proposition}
At $\tau=0.70,q=0.85$, the bound is 41 samples, versus 45 for the sequential design and 64 for the matched curtailed fixed design. The ordering is not universal: with $S=32$, $\tau=0.70$ gives $r=28$, attained false-declaration probability 0.019, and matched costs 24 sequential versus 30 curtailed fixed. At $\tau=0.90$, unanimity $r=32$ gives 39 versus about 30. For unanimity, \eqref{eq:curtail} reduces to $\E_q[N]=(1-q^S)/(1-q)$. In this particular $S=32$ family the ordering changes at $\tau=0.87$, the first threshold with $r=S$, not as a general theorem.

\subsection{Leader-declaration bound}\label{sec:leader-bound}
Let $c^\star$ be the vote-law leader with mass $\pi_{(1)}\ge\tau$, and aggregate all rival classes into a single outcome of mass $1-\pi_{(1)}$. For the monotone count rules considered here, if any individual rival reaches the declaration region, then the aggregate rival count also reaches it. Hence
\[
\Pr(\hat c\neq c^\star,D\mid\boldsymbol\pi)
\le \OC(1-\pi_{(1)}).
\]
Since $\Pr(\hat c=c^\star,D\mid\boldsymbol\pi)=\OC(\pi_{(1)})$, conditioning on a declaration gives
\[
\Pr(\hat c=c^\star\mid D,\boldsymbol\pi)
\ge
\frac{\OC(\pi_{(1)})}
{\OC(\pi_{(1)})+\OC(1-\pi_{(1)})}.
\]
The bound is tight when all nonleader vote mass is concentrated on a single rival.

\end{document}